\documentclass[11pt]{article}
\usepackage[preprint]{colm2026_conference}

\usepackage[T1]{fontenc}
\usepackage{mathpazo}

\usepackage{amsmath,amssymb,amsthm,mathtools}
\usepackage{booktabs}
\usepackage{graphicx}
\usepackage{microtype}
\usepackage{url}
\usepackage{hyperref}
\usepackage{algorithm}
\usepackage{algorithmic}

\definecolor{darkblue}{rgb}{0,0,0.5}
\hypersetup{colorlinks=true,citecolor=darkblue,linkcolor=darkblue,urlcolor=darkblue}

\theoremstyle{remark}

\theoremstyle{definition}

\title{LoRA Scaffolded Policy Optimization (LSPO): A Sampling-Time
Low-Rank Scaffold for Recovering Reinforcement-Learning Gradient on
Zero-Reward Cliff Prompts}

\author{Ken Ding \\
NVIDIA \\
\texttt{kennethd@nvidia.com}}

\begin{document}

\maketitle
\lhead{}

\begin{abstract}
Reinforcement learning from verifiable rewards (RLVR) for mathematical reasoning suffers from a structural blind spot: on ``cliff'' prompts---those on which every sampled rollout in a group fails---the group-normalized advantage is identically zero, so GRPO produces no gradient on precisely the prompts at the frontier of the model's capability. We introduce LoRA Scaffolded Policy Optimization (LSPO), a sampling-time mechanism that recovers this lost gradient. Each RL step, LSPO detects cliff prompts, fits a small low-rank (LoRA) adapter by a brief supervised step on their ground-truth solutions, re-rolls the cliffs with the base-plus-adapter model, splices the now-successful completions back into the RL batch with an importance-sampling correction, and takes a GRPO step on the base alone; the adapter receives only the supervised gradient and is discarded at checkpoint, yielding a base-only model. On DeepMath-103K with DeepSeek-R1-Distill-Qwen-1.5B, evaluated over $n{=}5$ paired seeds per arm at a matched $1000$-step reporting horizon, LSPO's $5$-seed mean matches or beats a DAPO baseline on \textbf{all $16$ (benchmark, \texttt{pass@}$k$) cells} ($15$ strict wins and one exact tie), with gains of up to $+10.7$ points on AIME24/\texttt{pass@}4, $+6.7$ points on AIME24 and AIME26 at \texttt{pass@}16, and $+2.4$ points on MATH500/\texttt{pass@}1; averaged over the $16$ cells the improvement is $+3.8$ points.
\end{abstract}

\section{Introduction}
\label{sec:intro}

Reinforcement learning from verifiable rewards (RLVR) has become the
dominant recipe for eliciting mathematical reasoning from large language
models \citep{Shao2024,DeepSeekAI2025}. Group Relative Policy
Optimization (GRPO) \citep{Shao2024} and its scaled-up successor DAPO
\citep{Yu2025dapo} dispense with a learned critic by sampling a group of
$K$ rollouts per prompt and normalizing each rollout's reward against the
group mean to form an advantage. This group-relative construction has a
structural blind spot. On a \emph{cliff} prompt---one where all $K$
sampled rollouts are wrong, so the group's total reward is zero---every
rollout carries the same reward, the within-group advantage is
identically zero, and the policy-gradient contribution of that prompt
vanishes exactly. The model therefore receives no learning signal from
precisely the prompts at the frontier of its capability: the problems it
cannot yet solve are the ones GRPO is
structurally unable to learn from. As training proceeds and the easy
prompts are mastered, the unsolved tail accumulates at this frontier,
and an increasing share of the batch produces no gradient.

Several lines of work attack this regime, and they cleave naturally into
two groups. The first targets the cliff directly, recovering the missing
signal from ground-truth solutions (Section~\ref{sec:relwork}). The second group combines
RL with a low-rank adapter,
though not aimed at the cliff problem specifically. Tina \citep{Tina2025}
trains a LoRA adapter \citep{Hu2021lora} by RL and ships the adapted
weights as the final model. BRIDGE \citep{Bridge2025} fuses supervised
and RL gradients in the base-model update and updates a LoRA teacher
using a cooperative-gain meta-objective derived from a bilevel
formulation.

We introduce \textbf{LoRA Scaffolded Policy Optimization (LSPO)}, a
sampling-time mechanism that wraps an existing policy-gradient RL loop
and recovers gradient on cliff prompts without modifying the loss. On
each step, LSPO detects the all-zero-reward cliffs, attaches a small
low-rank adapter and runs a brief supervised step on the cliff prompts'
ground-truth solutions (updating the adapter only, base frozen),
re-samples the cliffs with the adapter active to restore within-group
reward variance, then splices the now-successful rollouts back into the
RL batch and backpropagates the policy-gradient loss into the base
alone. What distinguishes LSPO from prior adapter-RL methods is three
design choices acting together: a \emph{sampling-time low-rank scaffold}
--- the adapter is a transient proposal distribution (never the
deliverable), kept low-rank and fine-tuned for only a handful of steps,
the intent being to keep the
importance-sampling correction well-conditioned rather than in the
high-variance regime of further-off-policy proposals;
\emph{strict gradient routing} via a two-optimizer split that sends the
supervised gradient to the adapter only and the RL gradient to the base
only (no bilevel or meta-objective coupling); and \emph{splice-and-discard},
in which the adapter is reused only as an importance-sampling proposal
for the spliced cliff rows and is stripped at checkpoint. The supervised
signal therefore reaches the base only indirectly, through
adapter-elicited rollouts, and the deliverable is a base-only model that
carries no adapter weights. Figure~\ref{fig:lspo} summarises one training
step.

Our contributions are as follows:

\begin{itemize}
  \item \textbf{Method.} We introduce LSPO, a sampling-time low-rank
  scaffold that recovers RL gradient on zero-reward cliff prompts via a
  two-optimizer split (adapter trained by supervision only, base trained
  by RL only) and a splice-and-discard pipeline with an
  importance-sampling correction, yielding a base-only deliverable
  (Section~\ref{sec:method}).

  \item \textbf{Headline empirical result.} On DeepMath-103K with
  DeepSeek-R1-Distill-Qwen-1.5B \citep{DeepSeekAI2025}, over $n{=}5$
  paired seeds per arm at a matched $1000$-step reporting horizon, the
  $5$-seed mean of LSPO (per-iteration variant) matches or beats a DAPO
  baseline on \textbf{all $16$} (benchmark, pass@$k$) cells---$15$
  strict wins and one exact tie---across MATH500 and AIME24/25/26. The
  largest gains are $+10.7$ points on AIME24/pass@4 and $+6.7$ points on
  both AIME24 and AIME26 at pass@16; averaged over the $16$ cells the
  improvement is $+3.8$ points. Notably LSPO improves pass@1 on every
  benchmark ($+1.3$ to $+4.7$ points), so the gain is not merely a
  broadening of the sampled distribution
  (Section~\ref{sec:experiments}).

  \item \textbf{Cliff-conversion efficacy.} We measure the mechanism
  directly: across the $5$ runs the scaffold converts ${\sim}43\%$ of
  otherwise-zero-gradient cliff groups into groups carrying usable
  policy-gradient signal, with a tight per-seed spread
  (Section~\ref{sec:experiments-cliff}).
\end{itemize}

\section{Related Work and Background}
\label{sec:related}

\subsection{RL with verifiable rewards: GRPO and the DAPO recipe}

Reinforcement learning with verifiable rewards (RLVR) trains a policy
$\pi_\theta$ to generate a solution $y$ to a prompt $x$ and assigns a
binary outcome reward $r(x,y)\in\{0,1\}$ that is $1$ iff a verifier
accepts the final answer \citep{DeepSeekAI2025}. Group Relative Policy
Optimization (GRPO) \citep{Shao2024} removes the value network used by
PPO \citep{Schulman2017} by drawing a group of $G$ rollouts
$\{y_1,\dots,y_G\}$ per prompt and normalizing rewards within the group
to form the advantage
$\widehat{A_i} = (r_i - \mu)/\sigma$, where $\mu$ and $\sigma$ are the
group reward mean and standard deviation. The policy is then updated with
the PPO-style clipped surrogate
\begin{equation}
\label{eq:grpo}
L_{\mathrm{GRPO}} = -\,\mathbb{E}\!\left[\min\!\Big(\rho_t \widehat{A},\;
\mathrm{clip}(\rho_t,\,1-\varepsilon_{\mathrm{low}},\,1+\varepsilon_{\mathrm{high}})\,\widehat{A}\Big)\right],
\qquad
\rho_t = \frac{\pi_\theta(a_t\mid s_t)}{\pi_{\mathrm{old}}(a_t\mid s_t)},
\end{equation}
where $\rho_t$ is the importance ratio between the current policy and the
sampling (behavior) policy.

\subsection{The cliff problem}
\label{sec:cliff}

The group-normalized advantage in Eq.~\eqref{eq:grpo} creates a sharp
failure mode. For a given prompt $x$ the outcome falls into one of three
cases: (i) all $G$ rollouts succeed, so every reward equals the group
mean and all advantages are zero; (ii) the rollouts are mixed, yielding
positive advantages for the successes and negative for the failures ---
the standard learning regime; or (iii) all $G$ rollouts fail, so the
group reward sum is zero, $\sigma$ is zero, and every within-group
advantage is identically zero. Case (iii) --- a \emph{cliff} prompt, the
``learning cliff'' of \citet{zhang2026scafgrpo} --- contributes \emph{no}
gradient to the policy update. These are precisely
the prompts at the frontier of the model's capability: too hard for any
of the $G$ samples to solve, yet exactly where a learning signal is most
needed. Under standard GRPO the cliff boundary can therefore only advance
indirectly, through weight sharing as the model learns on nearby
intermediate-difficulty prompts; there is no direct gradient on the
cliffs themselves.

\subsection{Low-rank adapters}

Low-Rank Adaptation (LoRA) \citep{Hu2021lora} freezes the pretrained weight
matrix $W_0$ and learns a low-rank update $\Delta W = BA$ with
$B\in\mathbb{R}^{d\times r}$, $A\in\mathbb{R}^{r\times k}$ and rank
$r \ll \min(d,k)$, so that the adapted forward pass computes
$(W_0 + BA)x$. Because only $A$ and $B$ are trainable, adaptation is
parameter-efficient and the adapter can be attached, detached, or reset
cheaply. LSPO exploits exactly this property: it uses a small,
short-lived LoRA adapter purely as a sampling-time mechanism, never
shipping it.

\subsection{Related work}
\label{sec:relwork}

\paragraph{Rescuing cliffs with a scaffold.} A cliff group can be
repaired in place, by synthesising a non-zero advantage through
entropy-modulated shaping or a confidence-weighted penalty on wrong
answers \citep{le2026rlzvp,feng2025lens}; this is far cheaper than LSPO,
but on an all-fail group it can only push probability mass \emph{away}
from the observed failures, never toward a success. Methods that instead
rescue the cliff---LSPO among them---re-sample the failed prompt under
privileged guidance and return the successes to the batch. The move
predates RLVR: STaR~\citep{zelikman2022star} recovered failed problems by
conditioning on the ground-truth answer, within a
generate--filter--finetune lineage that consumes the recovered trajectory
as a cross-entropy target rather than an RL sample. The closest of
these hints in the context and corrects the importance ratio back to the
hint-free prompt \citep{nath2025guide}. LSPO's scaffold is a weight
perturbation rather than a prompt, so its ratio compares two parameter
settings; and it splices only the verifier-passing rows, leaving the rest
of the group on-policy, where that method replaces the group wholesale.
The others differ mainly in how the hint is constructed---tiered from
abstract concepts to concrete steps \citep{zhang2026scafgrpo}, taken as a
prefix of the ground-truth trace \citep{liu2025ghpo,zhang2025bread},
generated by the policy itself
\citep{chen2026nurl,liao2026sage}, or produced by a separate hinter
policy \citep{xia2026hill}; all place the scaffold in the prompt, so
answer leakage into the sampled trajectory is a live concern, which for
LSPO it structurally is not.
LatentRevise also fits its scaffold rather than writing it: a soft prefix
in input-embedding space, optimised against the gold answer and then
discarded \citep{guo2026latentrevise}. The prefix is still an input, and
refitting it per prompt against that prompt's own answer is a more direct
privileged channel than LSPO's single low-rank update shared across the
whole cliff set.
\paragraph{Off-policy rows and privileged supervision.} POPO substitutes
a variance-bearing group drawn from a prioritised replay buffer under a
decoupled importance correction \citep{mao2026popo}; unlike replay, LSPO
can help on prompts the
policy has never solved. ZPPO argues that injecting a teacher's
response into the policy gradient breaks the on-policy assumption, and so
places that response in the prompt instead, as an anonymised candidate
the student must pick out \citep{lee2026zppo}. LSPO reaches the same
conclusion by another route: the ground-truth solution trains only the
adapter, and what enters the base's gradient is the model's own rollout,
importance-corrected. LUFFY imports off-policy
teacher traces into RLVR directly \citep{yan2025luffy}---the
further-off-policy regime that Section~\ref{sec:method} argues against on
variance grounds.

\paragraph{Where the supervision lands.} Hybrid distillation /
privileged self-distillation (HDPO) adds a student--teacher divergence
term to the RL objective, so that cliff prompts contribute a supervised
distillation signal optimised jointly with the policy gradient on the
model's own parameters \citep{Ding2026hdpo}; ReLIFT alternates phases of
full-model supervised fine-tuning on hard-prompt ground-truth solutions
with phases of RL, supplying cliff signal by overwriting the base weights
with expert traces \citep{Ma2025}. In both, a supervised loss is applied
to the base's own parameters---jointly with the policy gradient in the
first case, in alternating phases in the second. LSPO applies none: the
RL objective is left exactly as it was, the intervention acts only on the
proposal distribution, and the supervision terminates in an adapter that
is discarded. In BRIDGE neither signal is confined to one weight set: the base takes a
fused SFT-plus-RL gradient, and the adapter is updated under a bilevel
cooperative-gain objective \citep{Bridge2025}. LSPO routes the two to
disjoint parameter sets with no meta-objective, maintains two weight sets
rather than three, and activates only on cliff prompts. Where Tina's adapter is trained by RL
and shipped as the final model \citep{Tina2025}, LSPO's receives only
supervised gradient and is discarded.

\section{Method: LoRA Scaffolded Policy Optimization}\label{sec:method}

\begin{figure}[t]
\centering
\includegraphics[width=0.95\linewidth]{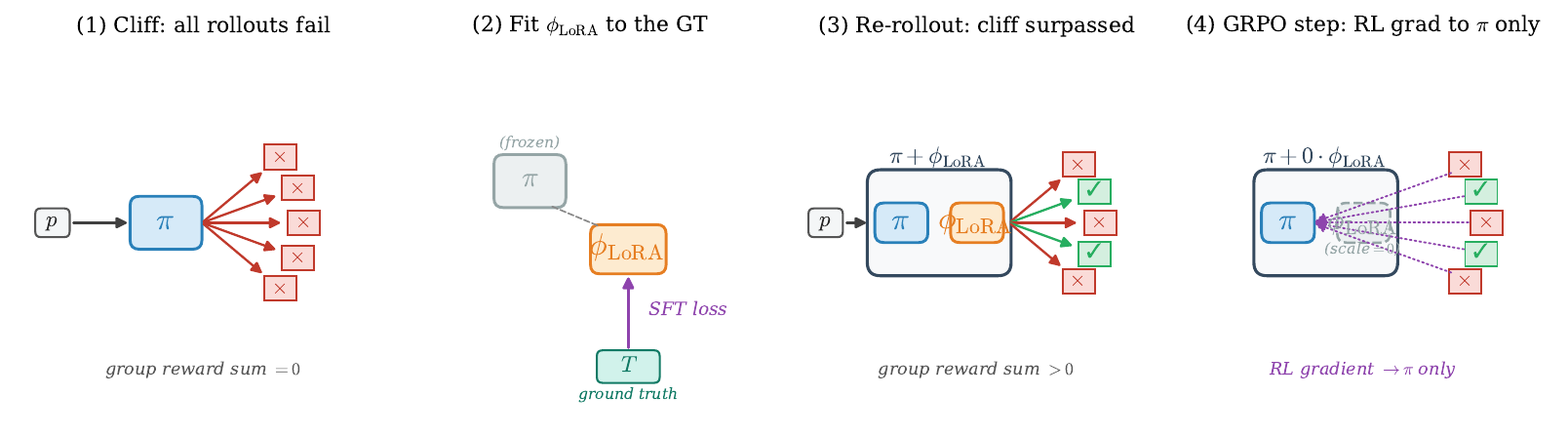}
\caption{The LSPO cliff-rescue mechanism. \textbf{(1)} On a cliff prompt
$p$, all $K$ rollouts from the base policy $\pi$ fail (red); the group
reward sum is zero and standard GRPO has no learning signal.
\textbf{(2)} A trainable LoRA adapter $\phi_\text{LoRA}$ is composed with
the (frozen) base $\pi$ and fitted via SFT on the dataset's ground-truth
trajectory $T$. \textbf{(3)} Under the composed policy
$\pi + \phi_\text{LoRA}$, some rollouts on the same prompt now pass the
verifier (green): the cliff is surpassed. Successful rollouts are spliced
into the GRPO batch as positive examples. \textbf{(4)} The GRPO step runs
on the same batch with the LoRA branch scale set to 0 (dashed grey), so
the RL gradient (purple, dotted) lands on $\pi$ only; $\phi_\text{LoRA}$
receives no RL gradient and is reset before the next step.}
\label{fig:lspo}
\end{figure}

LSPO is a sampling-time mechanism that wraps an existing policy-gradient RL loop
(here GRPO~\citep{Shao2024} with a DAPO-style recipe~\citep{Yu2025dapo}) and
recovers gradient on \emph{cliff} prompts---prompts on which every sampled
rollout in the group scores zero reward, so the group-relative advantage, and
hence the policy gradient on that prompt, is identically zero. The core idea is
to attach a small low-rank adapter~\citep{Hu2021lora} as a \emph{transient
scaffold}: it is supervised-fine-tuned on the cliff prompts' own ground-truth
solutions, used only to re-sample those prompts, and then discarded. Successful
adapter-elicited rollouts are spliced back into the RL batch so that the
within-group reward variance---and the RL gradient---is restored, while the
gradient that actually updates the deliverable flows into the base model alone.

\subsection{Algorithm}\label{sec:method-algo}

Figure~\ref{fig:lspo} illustrates the mechanism and
Algorithm~\ref{alg:lspo} states the per-step loop. Each RL step:
\textbf{(A)} sample $K$ completions per prompt from the base and flag as
cliffs those whose group scores zero total reward; \textbf{(B)} attach a
LoRA adapter and run a brief supervised step on the cliff prompts'
ground-truth solutions, base frozen; \textbf{(C)} re-sample the cliff
prompts with the adapter on; \textbf{(D)} swap each successful
re-rollout into its failed counterpart's slot, leaving every other row
untouched; \textbf{(E)} apply the RL update to the base alone and discard
the adapter.

The swap in (D) is deliberately minimal: a failed re-rollout carries no
signal the group does not already have from its base counterpart, so
admitting it would add off-policy mass without adding information.
Advantages are recomputed on the spliced batch, and each spliced row's
base-plus-adapter probability is retained for the importance correction
(Section~\ref{sec:method-is}).

\begin{algorithm}[t]
\caption{LoRA Scaffolded Policy Optimization (per-step loop)}
\label{alg:lspo}
\begin{algorithmic}[1]
\REQUIRE Base policy $\pi_\theta$, prompt set $\mathcal{X}$, ground truth
         $\{y^*\}$, reward $R$, rollouts per prompt $K$, clip bounds
         $(\varepsilon_{\mathrm{low}}, \varepsilon_{\mathrm{high}})$,
         RL optimizer $\mathrm{Opt}_{\mathrm{RL}}$ (updates $\theta$),
         SFT optimizer $\mathrm{Opt}_{\mathrm{SFT}}$ (updates adapter $\phi$)
\FOR{each RL training step}
  \STATE Sample batch $B \subset \mathcal{X}$; for each $x\in B$ draw
         $K$ rollouts $y^{(k)} \sim \pi_\theta(\cdot \mid x)$ and record per-row
         sampler density $\pi_{\mathrm{samp}} \leftarrow \pi_\theta$
  \STATE Score every rollout with $R$; \quad
         $\mathcal{C} \leftarrow \{x \in B : \textstyle\sum_k R(x,y^{(k)}) = 0\}$
         \quad\textit{// cliffs: zero RL gradient}
  \IF{$\mathcal{C} \neq \emptyset$}
    \STATE Attach low-rank adapter $\phi$ to $\pi_\theta$ (base frozen); \
           $\phi \leftarrow \mathrm{Opt}_{\mathrm{SFT}}$ on
           $\{(x, y^*) : x \in \mathcal{C}\}$
           \quad\textit{// supervised gradient to $\phi$ only}
    \STATE Re-sample $\mathcal{C}$ with adapter on:
           $\tilde{y} \sim \pi_{\theta+\phi}(\cdot \mid x)$; \
           $\mathcal{S} \leftarrow \{(x,\tilde{y}) : R(x,\tilde{y}) = 1\}$
    \STATE Splice $\mathcal{S}$ into $B$ in place of the failed cliff rollouts;
           for each $(x,\tilde y) \in \mathcal{S}$ overwrite
           $\pi_{\mathrm{samp}} \leftarrow \pi_{\theta+\phi}$
           \quad\textit{// IS denominator switches to base+adapter}
  \ENDIF
  \STATE Compute group-normalized advantages $\hat A_y$ on the (possibly spliced) batch $B$
  \STATE Per-token IS ratio: \quad
         $\rho_t(\theta) \;=\;
         \dfrac{\pi_\theta(y_t \mid x, y_{<t})}{\pi_{\mathrm{samp}}(y_t \mid x, y_{<t})}$
         \quad\textit{// $\pi_{\mathrm{samp}}$ equals $\pi_\theta$ for on-policy rows, $\pi_{\theta+\phi}$ for spliced cliffs}
  \STATE $\theta \leftarrow \mathrm{Opt}_{\mathrm{RL}}$ on the clipped surrogate
         (RL gradient to base only; adapter receives none):
         \[
         \mathcal{L}_{\mathrm{GRPO}}
         \,=\, -\frac{1}{T}\sum_{(x,y)\in B}\sum_{t}
         \min\!\big(
           \rho_t\, \hat A_y,\,
           \mathrm{clip}(\rho_t,\, 1{-}\varepsilon_{\mathrm{low}},\, 1{+}\varepsilon_{\mathrm{high}})\, \hat A_y
         \big)
         \]
         where $T = \sum_{(x,y)\in B} |y|$ is the total tokens in the batch
  \STATE Discard / reset adapter $\phi$
\ENDFOR
\STATE \textbf{return} base model $\pi_\theta$ \quad\textit{// adapter never shipped}
\end{algorithmic}
\end{algorithm}

\subsection{Gradient routing}\label{sec:method-routing}

The defining architectural choice of LSPO is that gradient routing is
\emph{strictly disjoint}. We maintain two optimizers. The supervised optimizer
$\mathrm{Opt}_{\mathrm{SFT}}$ updates the low-rank adapter only, on the
cross-entropy of the cliff ground-truth solutions (stage B). The RL optimizer
$\mathrm{Opt}_{\mathrm{RL}}$ updates the base only, on the GRPO surrogate over
the spliced batch (stage E). No parameter is touched by both objectives: the
base never receives supervised gradient, and the adapter never receives RL
gradient. The supervised signal reaches the base \emph{only indirectly}---through
the adapter-elicited rollouts that are spliced into the RL batch and then learned
by the ordinary policy gradient. Because a fresh adapter is fitted at every
step, that signal always measures the gap from the \emph{current} base to the
cliff solutions, with no adapter state carried across steps.

\subsection{Importance-sampling correction}\label{sec:method-is}

Spliced rows are off-policy: they were drawn from
$\pi_\mu := \pi_{\theta+\phi}$, not from the base $\pi_\theta$ whose
parameters the update modifies. The correction is a change of measure.
The quantity we want is the base's policy gradient, and
\begin{equation}
\mathbb{E}_{y\sim\pi_\theta}\!\left[A(y)\,\nabla_\theta \log \pi_\theta(y)\right]
=\;
\mathbb{E}_{y\sim\pi_\mu}\!\left[\frac{\pi_\theta(y)}{\pi_\mu(y)}\,A(y)\,\nabla_\theta \log \pi_\theta(y)\right],
\label{eq:cov}
\end{equation}
so each spliced row must carry the weight $\pi_\theta(y)/\pi_\mu(y)$,
which factorises per token into
$\prod_t \pi_\theta(y_t\mid y_{<t})/\pi_\mu(y_t\mid y_{<t})$.

The implementation instead uses the corresponding per-token ratio in the
clipped surrogate,
$\rho_t = \pi_\theta(y_t\mid y_{<t})/\pi_{\mathrm{prev}}(y_t\mid y_{<t})$
\citep{Schulman2017,Shao2024}. For ordinary rows $\pi_{\mathrm{prev}}$ is
the base; for spliced rows we set $\pi_{\mathrm{prev}} = \pi_\mu$, using
the base-plus-adapter log-probabilities recorded at stage~(D). No new
estimator is introduced --- the denominator is simply the distribution
that generated the row.

Keeping the adapter small keeps that ratio well-conditioned. This is a
variance motivation rather than an exactness
guarantee. Because the implementation selects verifier-passing proposal
rows and applies clipped per-token ratios rather than an unclipped
sequence-product weight, we do not claim that its update is an unbiased
sequence-level policy-gradient estimator or provide a convergence
guarantee; formal characterization is deferred to future work.

\section{Experiments}
\label{sec:experiments}

\subsection{Setup}
\label{sec:experiments-setup}

We evaluate LSPO on mathematical reasoning. The base model is
DeepSeek-R1-Distill-Qwen-1.5B \citep{DeepSeekAI2025}, a distilled reasoning
model. Training uses DeepMath-103K \citep{He2025DeepMath}, a public collection
of competition-style mathematics problems with verifiable final answers
(HuggingFace \texttt{zwhe99/DeepMath-103K}). We validate on four held-out
benchmarks: MATH500 \citep{Hendrycks2021,Lightman2024} and the AIME 2024, 2025,
and 2026 competition sets (AIME24/25/26),\footnote{We use the publicly
released problem sets from the Mathematical Association of America's
American Invitational Mathematics Examination for 2024, 2025, and 2026.
Problems and official answer keys are available via the MAA archive and
community wikis such as
\url{https://artofproblemsolving.com/wiki/index.php/AIME_Problems_and_Solutions}.}
reporting best pass@$k$ for
$k\in\{1,4,8,16\}$. Validation runs every 10 training steps. pass@$k$ is
computed by the first-$k$-slice estimator (whether any of the first $k$ of
the $16$ validation rollouts passes the verifier); it is unbiased for
pass@$k$ under exchangeability of the rollouts, though higher-variance
than the combinatorial estimator of \citet{Chen2021codex} which averages
over all size-$k$ subsets. The estimator is applied identically to both arms.

\paragraph{Seeds and reporting horizon.} Both arms are trained with the
same seed set $\{42,43,44,45,46\}$, and all analysis is paired seed-by-seed.
We fix a \textbf{1000-step reporting horizon}: runs are launched with a
1500-step budget, and every peak, comparison, and aggregate reported in
this paper is computed over validation events at step ${\leq}1000$,
identically for LSPO and DAPO. We do not characterize behavior beyond
this horizon in this version.

All training is run on $8\times$H100 GPUs. The base RL algorithm is GRPO
\citep{Shao2024} configured with the DAPO recipe
\citep{Yu2025dapo}: Clip-Higher with asymmetric ratio clipping
($\varepsilon_{\text{low}}{=}0.20$, $\varepsilon_{\text{high}}{=}0.28$),
token-level policy-gradient loss, and overlong-reward shaping (a soft length
penalty that engages past 7168 generated tokens up to the 8192-token
generation limit). Dynamic sampling is disabled. Each step samples 32 prompts
with 16 generations per prompt (512 rollouts/step). The base model is updated
with AdamW \citep{Loshchilov2019} at learning rate $1\mathrm{e}{-}6$; we apply
no reference-policy KL penalty.

The LSPO scaffold uses a LoRA \citep{Hu2021lora} adapter of rank 16
(\(\alpha{=}64\)) fine-tuned on cliff ground-truth solutions for
\texttt{sft\_steps\_per\_new\_cliff}${=}4$ steps with a separate AdamW
optimizer at learning rate \(2.5\mathrm{e}{-}5\). Cliffs are groups whose total
reward is zero; only successful adapter-elicited rollouts replace their failed
counterparts (partial swap), and spliced rows reuse the base-plus-adapter
log-probabilities as the importance-sampling denominator. The adapter receives
supervised gradient only and the base receives RL gradient only, per
Section~\ref{sec:method}. Unless stated otherwise, LSPO denotes the
per-iteration variant (a fresh adapter each step). Our \textbf{baseline} is
\emph{identical} in every respect---data, recipe, optimizer, hardware, and step
budget---with the LSPO scaffold disabled, i.e.\ plain DAPO-style GRPO. This
isolates the contribution of the scaffold.

\subsection{Main result: peak-vs-peak comparison}
\label{sec:experiments-main}

Table~\ref{tab:lspo-main} reports the full 16-cell peak comparison at the
matched reporting horizon (step ${\leq}1000$) with $n{=}5$ paired seeds for
both arms. For each (run, benchmark, $k$) we take the peak value over that
run's validation trace, then average across the five seeds. LSPO's mean
matches or beats DAPO's on \emph{every cell}, with \textbf{15 strict wins
and one exact tie} (AIME25/pass@8: $46.00$ vs.\ $46.00$; at integer counts
both arms recover $69$ of $150$ seed-problem pairs).

The advantage is largest on AIME24 (mean $+4.7$ to $+10.7$ points across
$k$, peaking at $+10.66$ points at pass@4) and on AIME26/pass@16
($+6.67$ points). MATH500 gains are smaller in absolute terms ($+0.4$ to
$+2.4$ points) because both arms approach the benchmark ceiling. Notably,
LSPO improves pass@1 on every benchmark ($+1.3$ to $+4.7$ points): the
improvement is not merely a broadening of the sampled distribution but a
shift in single-sample accuracy. Averaged over all $16$ cells, LSPO scores
$59.90$ against DAPO's $56.08$, a $+3.82$-point improvement.

Because peaks are taken per cell independently, the peak steps differ
across cells, so Table~\ref{tab:lspo-main} reports best-achievable-per-metric
during training rather than the performance of any single checkpoint.
Both arms are measured under identical protocol, so the paired comparison
is unaffected.

\begin{table}[t]
\centering
\small
\setlength{\tabcolsep}{4pt}
\begin{tabular}{llrrrrr}
\toprule
\textbf{Benchmark} & \textbf{pass@$k$} & \textbf{LSPO mean} & \textbf{LSPO range} & \textbf{DAPO mean} & \textbf{DAPO range} & \textbf{$\Delta$ (pts)} \\
& & (5 seeds) & & (5 seeds) & & \\
\midrule
MATH500 & pass@1  & \textbf{86.28} & 84.80--87.20 & 83.88 & 83.20--84.60 & $+2.40$ \\
MATH500 & pass@4  & \textbf{94.80} & 94.60--95.20 & 93.36 & 93.00--93.60 & $+1.44$ \\
MATH500 & pass@8  & \textbf{96.44} & 96.00--96.60 & 95.56 & 95.40--96.00 & $+0.88$ \\
MATH500 & pass@16 & \textbf{97.52} & 97.20--97.60 & 97.08 & 96.80--97.20 & $+0.44$ \\
\midrule
AIME24 & pass@1  & \textbf{37.33} & 33.33--40.00 & 32.66 & 30.00--33.33 & $\mathbf{+4.67}$ \\
AIME24 & pass@4  & \textbf{59.33} & 53.33--63.33 & 48.67 & 46.67--53.33 & $\mathbf{+10.66}$ \\
AIME24 & pass@8  & \textbf{65.33} & 63.33--70.00 & 56.67 & 56.67--56.67 & $\mathbf{+8.66}$ \\
AIME24 & pass@16 & \textbf{72.00} & 70.00--73.33 & 65.33 & 63.33--70.00 & $\mathbf{+6.67}$ \\
\midrule
AIME25 & pass@1  & \textbf{28.67} & 26.67--30.00 & 27.34 & 26.67--30.00 & $+1.33$ \\
AIME25 & pass@4  & \textbf{42.00} & 40.00--43.33 & 38.00 & 36.67--43.33 & $\mathbf{+4.00}$ \\
AIME25 & pass@8  & 46.00 & 43.33--50.00 & 46.00 & 43.33--46.67 & $+0.00$ \\
AIME25 & pass@16 & \textbf{54.00} & 50.00--56.67 & 51.33 & 50.00--53.33 & $+2.67$ \\
\midrule
AIME26 & pass@1  & \textbf{30.00} & 26.67--33.33 & 25.33 & 23.33--26.67 & $\mathbf{+4.67}$ \\
AIME26 & pass@4  & \textbf{42.66} & 40.00--43.33 & 40.00 & 36.67--43.33 & $+2.66$ \\
AIME26 & pass@8  & \textbf{49.33} & 46.67--53.33 & 46.00 & 43.33--50.00 & $+3.33$ \\
AIME26 & pass@16 & \textbf{56.67} & 53.33--60.00 & 50.00 & 46.67--53.33 & $\mathbf{+6.67}$ \\
\midrule
\multicolumn{2}{l}{\textit{Mean over 16 cells}} & \textbf{59.90} & --- & 56.08 & --- & $+3.82$ \\
\bottomrule
\end{tabular}
\caption{Peak pass@$k$ at the matched $1000$-step reporting horizon,
averaged across $n{=}5$ paired seeds per arm, on
DeepSeek-R1-Distill-Qwen-1.5B trained with DeepMath-103K ($8\times$H100).
LSPO is the per-iteration variant; DAPO is the identical recipe with the
scaffold disabled. LSPO's mean matches or beats DAPO's on all $16$ cells
($15$ strict wins $+$ $1$ exact tie at AIME25/pass@8). $\Delta$ entries
$\geq{+}4$ points are highlighted in the final column. Peak steps vary by
cell. AIME sets have $N{=}30$ problems (one problem $\approx 3.33$ points),
so single-cell deltas are coarse; the pattern across cells is the reliable
signal.}
\label{tab:lspo-main}
\end{table}

\subsection{Cliff-conversion efficacy}
\label{sec:experiments-cliff}

Table~\ref{tab:lspo-main} measures the outcome; this section measures the
mechanism. Across the $5$ LSPO runs at step ${\leq}1000$, $10{,}864$ cliff
groups were detected in total. The scaffold
--- four iterations of LoRA fitting on the cliff ground-truth solutions
followed by re-sampling --- surpassed approximately \textbf{$43\%$} of
them, i.e.\ produced at least one verifier-passing rollout on $4{,}680$
groups that were originally $0$-of-$K$ failures and therefore carried
exactly zero policy gradient. The per-seed surpass rate is tight
($41.3$--$45.5\%$, standard deviation $1.5\%$).

Two details sharpen the interpretation. First, the per-rollout post-scaffold
pass rate is only $5.8\%$ (cliff-weighted mean across seeds), so the
group-level $43\%$ predominantly reflects \emph{one} of the $K{=}16$
re-sampled rollouts succeeding rather than confident solving --- which is
all GRPO requires, since a single success restores within-group reward
variance. Second, the per-step surpass rate is bimodal ($31\%$ of
cliff-bearing steps have rate $0.0$; $21\%$ have rate $1.0$), so the
aggregate $43\%$ is a mean of extremes rather than a typical value.
By construction the DAPO baseline converts $0\%$ of its cliffs: the
leave-one-out advantage zeroes those groups' gradients outright.

\section{Limitations}
\label{sec:limitations}

\paragraph{Scope.} We evaluate one model
(DeepSeek-R1-Distill-Qwen-1.5B), one dataset (DeepMath-103K), one recipe
family (GRPO with the DAPO configuration), and one scaffold configuration
(per-iteration variant, LoRA rank $16$, four SFT steps). The scaffold's
behavior plausibly depends on adapter rank, SFT step count, splice
variant, and the size of the dataset's answer space --- a small-integer
answer space makes verifier-accepted guessing more likely than a large
one --- and we do not establish how the results transfer across these
axes.

\paragraph{Ground-truth requirement.} LSPO requires ground-truth
solutions (not merely final answers) for the cliff prompts, since the
scaffold is fit by supervised learning on those derivations. This is
available in DeepMath-103K but restricts applicability to datasets that
ship worked solutions.

\section{Conclusion}
\label{sec:conclusion}

We introduced LSPO (LoRA Scaffolded Policy Optimization), a sampling-time
mechanism that recovers reinforcement-learning gradient on zero-reward
``cliff'' prompts---those frontier prompts where every sampled rollout fails,
the within-group advantage is identically zero, and standard GRPO therefore
provides no learning signal. LSPO attaches a small low-rank adapter, fine-tunes
it only on the cliff prompts' ground-truth solutions, re-samples the cliffs
with the adapter active to restore within-group reward variance, and splices
the resulting successes back into the RL batch. Two design choices keep the
deliverable a clean base-only model: strictly disjoint gradient routing via a
two-optimizer split (the supervised signal reaches the adapter only and the RL
gradient reaches the base only, reusing the standard policy-ratio term as the
importance-sampling correction for the spliced rows), and a splice-and-discard
policy that strips the adapter at checkpoint so it is never shipped. Over
$n{=}5$ paired seeds at a matched $1000$-step reporting horizon, LSPO's
mean matches or beats a DAPO baseline on all $16$ peak pass@$k$ cells
($15$ strict wins and one exact tie), with gains up to $+10.7$ points on
AIME24/pass@4 and $+3.8$ points averaged across the $16$ cells, and with
improvements on pass@1 for every benchmark. The mechanism is directly
measurable: roughly $43\%$ of otherwise-zero-gradient cliff groups are
converted into groups that carry usable policy-gradient signal. We view
LSPO as evidence that the cliff problem
can be addressed at sampling time, with the supervision localized in
transient low-rank parameters rather than written irreversibly into the
base.

\section*{LLM Usage Statement}

The research idea underlying LSPO---using a transient, sampling-time low-rank
scaffold to recover reinforcement-learning gradient on zero-reward cliff
prompts---originated with the author. However, an AI language model (Claude,
Anthropic) was used extensively throughout this project in ways that go beyond
minor writing assistance. Specifically: (1)~the paper text, including the
related-work survey, the method exposition, and the mechanism-analysis and
discussion sections, was substantially drafted and edited with LLM assistance;
(2)~the LLM was used as a research collaborator to stress-test explanations
for the observed behavior, to assemble and check the related-work survey, and
to position LSPO against prior methods; a second model (OpenAI Codex, GPT) was
used as an independent verifier, recomputing every reported number from the
raw training logs and auditing citations; and
(3)~the importance-sampling correction described in
Section~\ref{sec:method-is} --- reusing the standard PPO/GRPO policy-ratio
term with the base-plus-adapter sampling log-probabilities as the ratio's
denominator on spliced cliff rows --- was proposed by the LLM during the
method-design phase and adopted by the author.
All experimental results (training
runs, metric measurements, and rollout dumps) were produced by
the author without LLM involvement.

\bibliography{lspo_colm2026}
\bibliographystyle{colm2026_conference}

\end{document}